\documentclass[letterpaper, 10 pt, conference]{ieeeconf}  

\IEEEoverridecommandlockouts                              

\usepackage{graphicx} 
\usepackage[ruled,vlined]{algorithm2e}
\usepackage{mathtools}
\DeclarePairedDelimiterX{\norm}[1]{\lVert}{\rVert}{#1}
\usepackage{amsmath}
\usepackage{multirow}
\usepackage{xcolor}

\title{\LARGE \bf
Fast User Adaptation for Human Motion Prediction in Physical Human--Robot Interaction
}

\author{Hee-Seung Moon and Jiwon Seo
\thanks{This work was supported in part by the Basic Science Research Program through the National Research Foundation of Korea (NRF) funded by the Ministry of Education (NRF-2018R1D1A1B07043580) and in part by the Institute of Information and Communications Technology Planning and Evaluation (IITP) grant funded by the Korea government (KNPA) (2019-0-01291).
\textit{(Corresponding author: Jiwon Seo.)}}
\thanks{The authors are with the School of Integrated Technology, Yonsei University, Incheon 21983, Korea (e-mail: hs.moon@yonsei.ac.kr; \mbox{jiwon}.seo@yonsei.ac.kr).}%
}

\begin{document}
\maketitle
\thispagestyle{empty}
\pagestyle{empty}


\begin{abstract}

Accurate prediction of human movements is required to enhance the efficiency of physical human--robot interaction. Behavioral differences across various users are crucial factors that limit the prediction of human motion. Although recent neural network-based modeling methods have improved their prediction accuracy, most did not consider an effective adaptations to different users, thereby employing the same model parameters for all users. To deal with this insufficiently addressed challenge, we introduce a meta-learning framework to facilitate the rapid adaptation of the model to unseen users. In this study, we propose a model structure and a meta-learning algorithm specialized to enable fast user adaptation in predicting human movements in cooperative situations with robots. The proposed prediction model comprises shared and adaptive parameters, each addressing the user's general and individual movements. Using only a small amount of data from an individual user, the adaptive parameters are adjusted to enable user-specific prediction through a two-step process: initialization via a separate network and adaptation via a few gradient steps. Regarding the motion dataset that has 20 users collaborating with a robotic device, the proposed method outperforms existing meta-learning and non-meta-learning baselines in predicting the movements of unseen users.

\end{abstract}


\section{Introduction}

With recent advancements in robotics, collaborative robots are now considered to move in physical contact with humans~\cite{medina2011experience, mortl2012role, unhelkar2018human, moon2020sample}. One representative example of the physical human--robot interaction (pHRI) is a situation in which a person performs a task while receiving physical assistance from a robot~\cite{abbink2012haptic}. Under robotic guidance, humans can utilize the repeatability and accuracy of the robot, which leads to improved productivity and reduced workload~\cite{salvine2011benefits}. The movement of robots for human assistance can be planned based on human behavior prediction~\cite{medina2015synthesizing, moon2021optimal}. However, if the robot mispredicts the next human motion and a conflict between the human's intention and the robotic guidance occurs, it can lead to a decrease in the collaborative task performance and an increase in the discomfort of the human operator~\cite{passenberg2013exploring}. Therefore, expanding the robot's capability to predict human motion is garnering considerable interest in the HRI field~\cite{kanazawa2019adaptive, cheng2020towards}.

The different behavioral patterns of individual human operators (i.e., users) are contributing factors that limit the prediction of human motion. People have different behaviors owing to a variety of factors, such as their motor skills or personal preferences~\cite{mitsunaga2008adapting}. Recent deep learning-based approaches have succeeded in enhancing the accuracy of human motion prediction; however, only a few studies have addressed how to respond to different users. Most of the previous methods employed a neural network model with fixed parameters to predict the movements of various users. A straightforward alternative that can cope with various users is to train a new model from scratch or to fine-tune a pretrained model for each new user. However, acquiring sufficient data for every new user is time-consuming and impractical in real-world applications. Therefore, training a single prediction model with user-adaptive characteristics is crucial for further advancement in human motion prediction.

\begin{figure}[t]
  \centering
  \includegraphics[width=\linewidth]{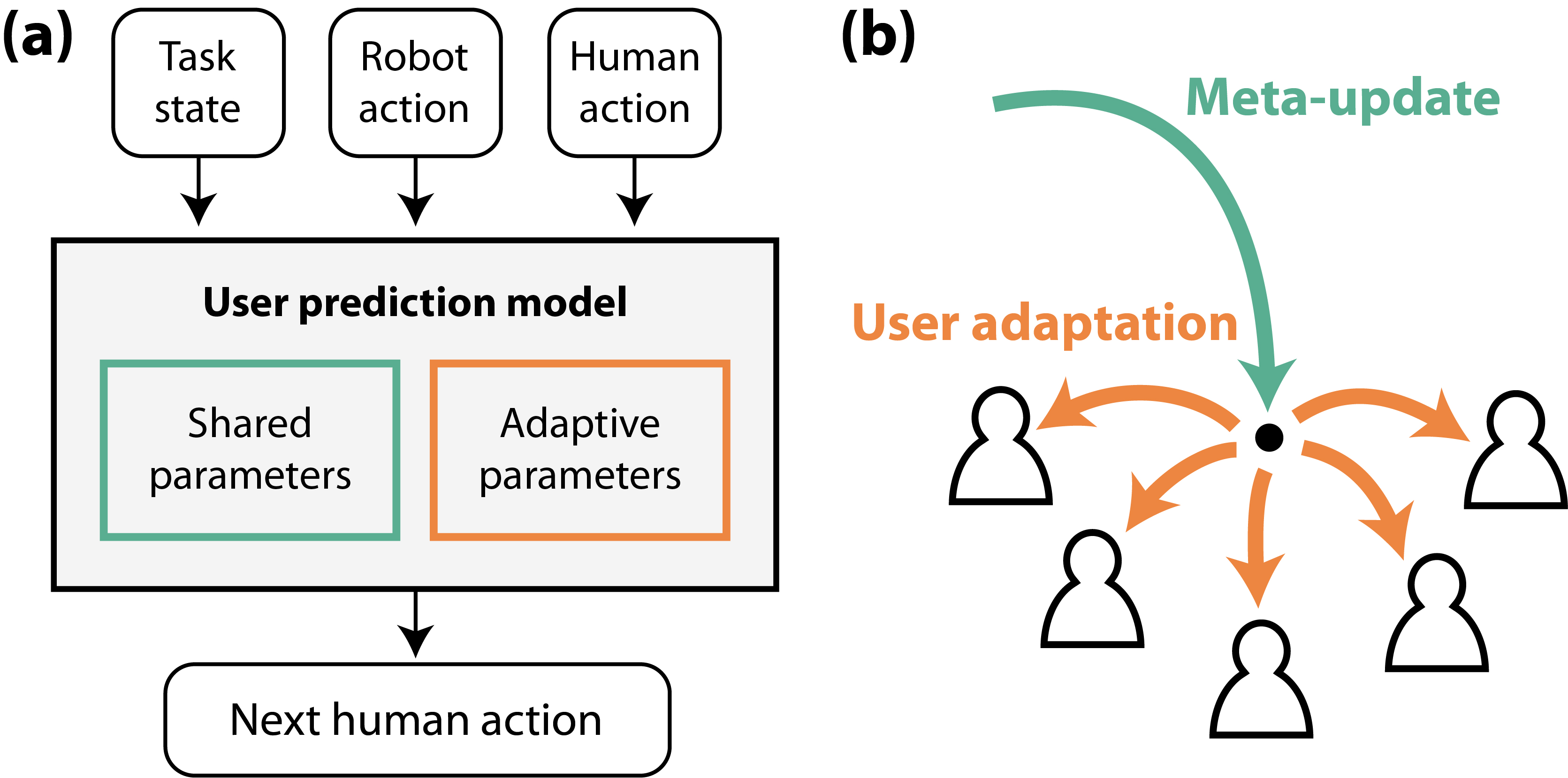}
  \caption{Overview of the proposed fast user adaptation approach. (a) Model comprising shared and adaptive parameters predicts the next human action based on current knowledge. (b) The shared parameters, which are trained via meta-update, handle general movements across the users, and the adaptive parameters, which are updated according to each user, address the individual differences.}
  \label{fig:01}
\end{figure}

To address this challenge, a \textit{meta-learning} approach can be considered as a possible solution. Meta-learning, \textit{learning to learn}, is a promising machine learning technique for solving the \textit{fast adaptation} problem, i.e., enabling a model to rapidly adapt to previously unseen tasks with small amounts of data. In the wake of \textit{model-agnostic meta-learning} (MAML)~\cite{finn2017model}, optimization-based meta-learning algorithms have demonstrated significant success in fast adaptation problems, such as few-shot image classification tasks.

The meta-learning approach is effective for solving fast adaptation problems. However, it is unclear whether the meta-learning approach is effective in the fast \textit{user} adaptation problem, that is, enabling a single model to swiftly adapt to previously unseen users with a small amount of their behavioral data. Adapting to different user behaviors in a cooperative situation with a robot has different characteristics from the problems previously addressed by the meta-learning approach, because of its unique situation. For example, the movements of users to perform a specific task can be divided into general movements (i.e., with low variance between users) to achieve a goal and individual movements (i.e., with high variance between users) affected by individual factors. Therefore, it is necessary to distinguish between them to adapt effectively  to different user behaviors. This distinction has not been the focus of previous applications using the meta-learning approach.

We have recently demonstrated in \cite{moon2021optimal} that the parameter adaptation of the user prediction model using MAML is effective in providing personalized haptic guidance to each user. This was the first attempt to apply meta-learning to a fast user adaptation problem. However, that study was limited to the direct application of MAML, which is not specifically designed to solve fast user adaptation problems. Therefore, the applicability of various meta-learning algorithms to fast user adaptation still remains unclear.

In this study, we demonstrate the feasibility of various existing meta-learning algorithms for solving fast user adaptation problems. Moreover, we propose a model structure and a meta-learning algorithm that is specialized for fast user adaptation. We focus on dividing human motions into common user movements and additional movements triggered by individual differences. Therefore, the proposed prediction model consists of shared and adaptive parameters, each of which is responsible for general and individual movements, respectively (Fig.~\ref{fig:01}(a)). In particular, the proposed method has two unique approaches: determining the user-specific initialization of the adaptive parameters via a separate network and enforcing the adaptive parameters to handle individual differences via a meta-loss function.

We evaluated the human motion prediction performance of the proposed method and compared it with several major meta-learning methods using a dataset acquired in a situation wherein a user performed a task while being assisted by a robotic device. The meta-learning methods exhibited better prediction performance than the non-meta-learning methods. This implies that meta-learning methods can be applied to solve fast user adaptation problems. In particular, our proposed meta-learning method with the initialization network and meta-loss function exhibited the best prediction accuracy among the meta-learning algorithms. In addition, we analyzed how the proposed model distinguished different users by visualizing the adaptive parameters that were adjusted to different users.

The contributions of this study are presented as follows: (1) We validate the applicability of existing meta-learning algorithms for fast user adaptation. To the best of our knowledge, this is the first attempt to compare the performance of meta-learning methods in solving fast user adaptation problems. (2) We propose a novel model structure and meta-learning algorithm specialized for enabling fast user adaptation in predicting human movements during pHRI. (3) We experimentally demonstrate that the proposed meta-learning method with our initialization network and meta-loss function exhibits the best accuracy compared to other meta-learning algorithms for predicting user movements during pHRI.

\section{Related Work}

\subsection{Human Motion Prediction}

Early studies on human motion prediction were developed based on probabilistic models such as the hidden Markov model~\cite{kulic2011incremental, power2015cooperative}, Gaussian mixture regression~\cite{kanazawa2019adaptive}, and probabilistic movement primitives~\cite{paraschos2013probabilistic}. The recent use of deep neural network architectures has resulted in a remarkable improvement in prediction performance. Fragkiadaki \textit{et al.}~\cite{fragkiadaki2015recurrent} first proposed an encoder-recurrent-decoder model structure, which successfully predicted human body movements using a dataset spanning multiple subjects and activity domains. Subsequent studies have focused on utilizing context information as clues for prediction, such as the motion data of nearby people~\cite{adeli2020socially}, or multimodal responses of the user~\cite{moon2019prediction}. In addition, there have been attempts to consider other learning techniques that increase prediction accuracy and robustness. For example, variational autoencoder~\cite{butepage2018anticipating} or adversarial learning~\cite{gui2018teaching, yasar2021scalable} frameworks have been adopted in motion prediction.

We focus on a learning framework that enables the proposed model to effectively respond to differences between individuals for human motion prediction, which has not been addressed in the aforementioned studies that employed fixed model parameters for all users. To solve this problem, we present a meta-learning approach that can quickly adapt the prediction model to novel users.

There have been attempts to apply a meta-learning technique to adapt the prediction model to novel tasks. Proactive and adaptive meta-learning (PAML)~\cite{gui2018few} integrates MAML and model regression networks~\cite{wang2016learning} to learn an effective adaptation strategy. MoPredNet~\cite{zang2020few} is a method with a parameter generation module that utilizes external memory. These two previous studies~\cite{gui2018few, zang2020few} focused on predicting human motion across specific categories (e.g., walking, eating, or smoking) without any interaction with a robot. 

Our work is distinguished from the prior works in that we consider a situation wherein a human and a robot physically interact. In the pHRI situation of our work (i.e., virtual air hockey environment, described in Section~\ref{sec:4-A}), the user behavior is strongly affected by the robotic guidance at every timestep, making it difficult to predict the user behavior over a long time horizon because the interacting robot's guidance over the horizon is already unpredictable. The robotic guidance depends on the opponent's play, which is obviously unpredictable. In~\cite{moon2021optimal}, it has been shown that even one-step human motion prediction can improve the user's task performance under a certain pHRI situation. Therefore, we focus on predicting the user's movement at the immediate next timestep given the dynamically changing and unpredictable robotic guidance of the current timestep.
It is worth mentioning that we have tested and compared the prediction performance when using the data of the past several timesteps as the input and when using the data of the current timestep only, but there was no significant performance difference.
Therefore, we decided to use the knowledge at the current timestep for the prediction.

\subsection{Optimization-based Meta-learning}

Optimization-based meta-learning is a technique that allows a model to learn a new task quickly via an optimization procedure based on small data samples. A powerful and representative example is MAML~\cite{finn2017model}, a meta-learning algorithm with a dual-structured training procedure consisting of inner and outer loop updates. The training procedure for MAML aims to attain model parameters that can reach task-specific parameters of a new task within a few gradient steps. Since MAML outperformed previous methods, such as the meta-learner with recurrent layers~\cite{ravi2017optimization}, various other optimization-based algorithms have followed, for example, Reptile~\cite{nichol2018first}, which simplifies the second-order gradient computation of the MAML; LEO~\cite{rusu2019meta}, which performs adaptation in the low-dimensional embedding of model parameters; and multimodal MAML~\cite{vuorio2019multimodal}, which pursues a more diverse task distribution through parameter modulation.

The aforementioned meta-learning algorithms updated all of the parameters of every layer during the adaptation process. However, Raghu \textit{et al.}~\cite{raghu2019rapid} discovered that the entire parameter adaptation changed the parameters of the last layer. Therefore, the authors proposed the ANIL algorithm that adapts only the parameters of the last layer while fixing the parameters of the body layers. Even with this simplification, ANIL exhibited a performance comparable to that of the other meta-learning algorithms utilizing the entire parameter adaptation. CAVIA~\cite{zintgraf2019fast} separates adaptive parameters (which are updated during the adaptation and condition the body layers) and shared parameters (which are fixed). By separating these parameters, CAVIA outperformed MAML despite adapting fewer parameters.

Fast user adaption can be a new application of existing meta-learning algorithms. However, the applicability of the existing meta-learning algorithms and a comparison of their performance have not yet been studied. We propose a novel meta-learning framework specialized in user movement prediction after investigating the user prediction performance of existing meta-learning algorithms. Inspired by the CAVIA model, our prediction model consists of shared and adaptive parameters. Shared and adaptive parameters predict general and individual user movements, respectively. Our model responds to movement differences between individual users by adjusting adaptive parameters. In the CAVIA algorithm, the adaptive parameters are updated for each new task from a zero vector. This initialization method can impede the ability to adapt to various tasks within a few gradient steps. Hence, we propose a model structure that determines the effective user-specific initialization of adaptive parameters. In addition, we suggest a meta-loss function for the meta-update that induces the adaptive parameters to handle individual differences exclusively.

\section{Proposed Method}

\subsection{Problem Definition}

The goal of our meta-learning approach is to train a human motion prediction model that can quickly adapt to a new user with small data samples. In a situation wherein a user performs a task while being guided by a robot, the prediction problem can be formulated as follows: to predict \(\mathbf{y}\), which indicates the human action at the next timestep, when given an input \(\mathbf{x}=\{\mathbf{x}_s, \mathbf{x}_r, \mathbf{x}_h\}\), which indicates knowledge at the current timestep consisting of \(\mathbf{x}_s\), interaction state (e.g., state of the cooperative task), \(\mathbf{x}_r\), robot action, and \(\mathbf{x}_h\), human action. We let \(\mathcal{U}_i\) denote the dataset collected from one user of index \(i\). \(\mathcal{U}_i\) consists of the interaction data of the \(N_i\) timestep length (i.e., \((\mathbf{x}, \mathbf{y})_i\) pairs of size \(N_i\)).

For the fast user adaptation problem, two batches \(\mathcal{D}^{spt}_i\) and \(\mathcal{D}^{qry}_i\) are given, each consisting of \((\mathbf{x}, \mathbf{y})_i\) pairs of size \(B\) (\(B \ll N_i\)) and sampled from the same dataset \(\mathcal{U}_i\) without overlapping. \(\mathcal{D}^{spt}_i\) is employed to adapt the parameters of the prediction model \(f\) to user \(i\), and the accuracy of the \(\mathbf{x} \rightarrow \hat{\mathbf{y}}\) mapping of the adapted model \(f^\prime\) for \(\mathcal{D}^{qry}_i\) is investigated. For the model training, a meta-dataset \(\mathcal{M}^{tr}\) is employed, which is composed of datasets spanning multiple users (e.g., \(\{\mathcal{U}_1, \mathcal{U}_2, ..., \mathcal{U}_m\}\)). Another meta-dataset \(\mathcal{M}^{test}\), collected from users who do not overlap with the users of \(\mathcal{M}^{tr}\), is adopted to evaluate the performance of the fast user adaptation on previously unseen users.

\subsection{Model Overview}
\label{sec:3-B}

\begin{figure}[t]
  \centering
  \includegraphics[width=\linewidth]{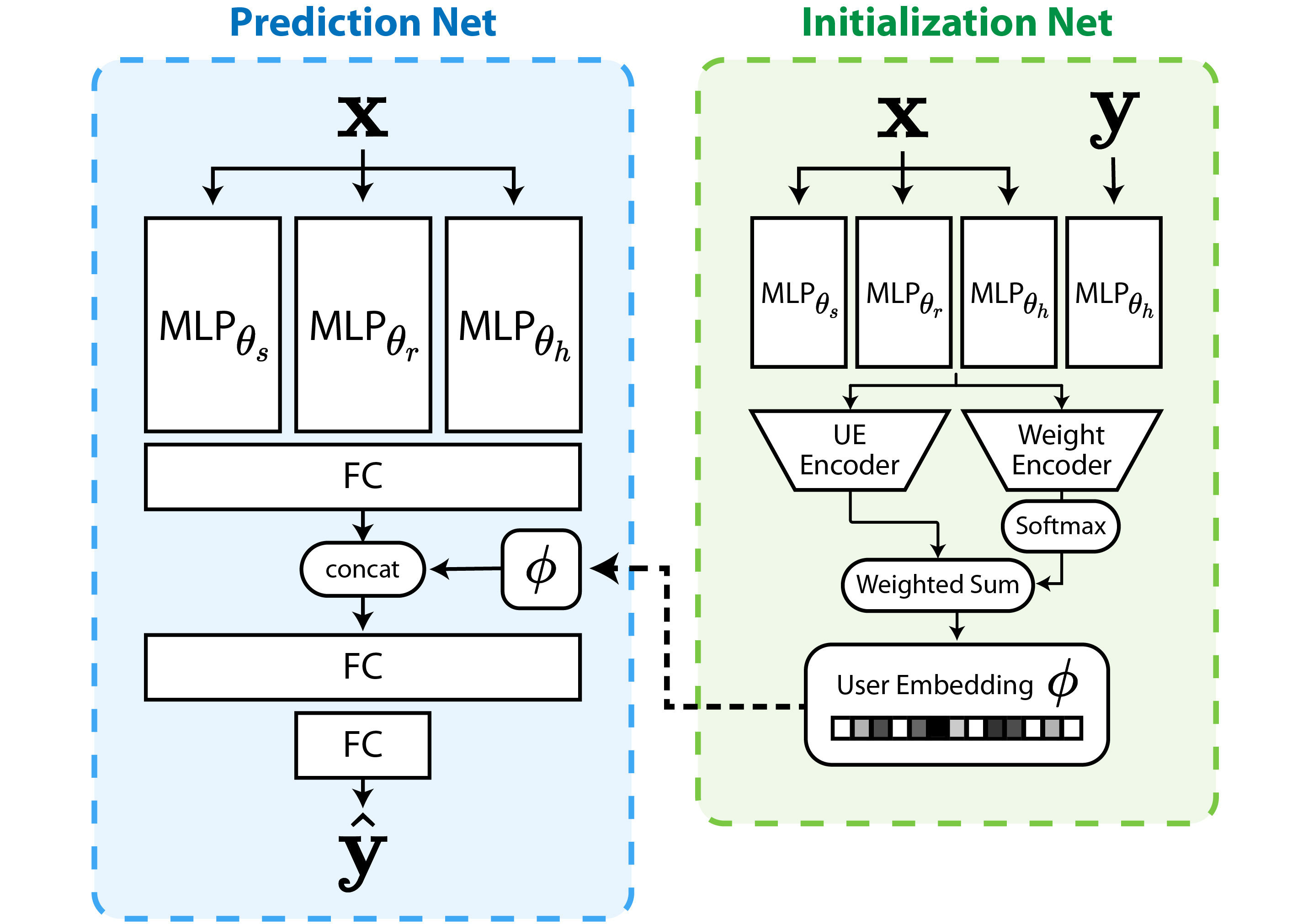}
  \caption{Overview of the proposed user prediction model structure.}
  \label{fig:02}
\end{figure}

Our model parameters consist of \(\theta\), which is shared across all users, and \(\phi\), which is adapted according to each user. Because \(\phi\) exhibits user-specific values after a series of initialization and adaptation processes, we refer to \(\phi\) as the \textit{user embedding} (UE). To implement a prediction model conditioned by \(\phi\), we concatenated \(\phi\) to a hidden state vector of the model’s middle layer and adopted the concatenated vector as the input of the next layer, which was originally proposed in~\cite{zintgraf2019fast}. In addition, we propose a model structure that determines the effective initial values of user embedding \(\phi_i\) for a user \(i\) based on small data samples \(\mathcal{D}^{spt}_i\). Accordingly, as illustrated in Fig.~\ref{fig:02}, the entire structure of our model is composed of two parts: a prediction network that outputs user-specific predicted movements of the user, and an initialization network that outputs the effective initial point of user embedding before adaptation. The shared parameters \(\theta\) contains all trainable parameters of the prediction network and the initialization network, except for the user embedding \(\phi\) (see Fig.~\ref{fig:03}(c)).

In the prediction network, three types of inputs (i.e., \(\mathbf{x}_s, \mathbf{x}_r,\) and \(\mathbf{x}_h\)) were fed into separate multilayer perceptron (MLP) blocks (with parameters \(\theta_s, \theta_r\), and \(\theta_h\), which are subsets of \(\theta\)) that extract each feature. Subsequently, feature vectors passed through the integrating layers to produce \(\hat{\mathbf{y}}\), which is the predicted user movement. The user embedding \(\phi\) was concatenated with the input vector of the second integrating layer to allow user-specific prediction.

The initialization network received the entire \(\mathcal{D}^{spt}_i\) consisting of \(B\) pairs of \((\mathbf{x}^{spt}, \mathbf{y}^{spt})_i\) as the input. To extract the features of the \(\mathbf{x}^{spt}\), MLPs that shared the parameters \(\theta_s, \theta_r,\) and \(\theta_h\) from the prediction network were employed, and an MLP with parameter \(\theta_h\) was again adopted to extract the features of \(\mathbf{y}^{spt}\). Note that \(\mathbf{x}_h\) and \(\mathbf{y}\) are vectors of the same format that represent human actions. To obtain a representative \(\phi_i\) corresponding to \(B\) pairs of \((\mathbf{x}^{spt}, \mathbf{y}^{spt})_i\), we first obtained the user embedding candidates and weight values, each corresponding to one \((\mathbf{x}^{spt}, \mathbf{y}^{spt})_i\) pair, by feeding the feature vectors into a \textit{UE encoder} and \textit{weight encoder}, respectively. The weight value indicates the extent to which the corresponding pair expresses the user characteristics. The relative importance of the corresponding pair among the batches was determined by passing the weight value through the softmax function. Therefore, the representative \(\phi_i\) (\(1 \times S\), where \(S\) is the size of \(\phi\)) was acquired by matrix multiplication of the weight values (\(1 \times B\)) and \(B\) user embedding candidates (\(B \times S\)). We employed three-layered MLPs for each UE and weight encoder. For the UE encoder, because it is unnecessary to embed different users to have the same bias, we deleted the bias term of the last layer.

\subsection{Meta-learning Procedure}

\begin{algorithm}[t]
\SetKwInput{KwRequire}{Require}
\SetAlgoLined
\DontPrintSemicolon
\KwRequire{Meta-dataset \(\mathcal{M}^{tr}\)}
\KwRequire{Hyper-parameters \(\alpha, \beta, d\)}
  Randomly initialize \(\theta\)\;
  \While{max iteration}{
    Sample users \(\mathbf{U}=\{\mathcal{U}_i\}\) where \(\mathcal{U}_i \sim \mathcal{M}^{tr}\)\;
    \For{all \(\mathcal{U}_i \in \mathbf{U}\)}{
      Sample \(\mathcal{D}^{spt}_i, \mathcal{D}^{qry}_i\) from \(\mathcal{U}_i\)\;
      Initialize \(\phi_i\) using \(\mathcal{D}^{spt}_i\) \textbf{(UE initialization)}\;
      Initialize \(\phi^\prime_i = \phi_i, \alpha^\prime = \alpha\)\;
      \For{number of adaptation steps}{
        Compute regression loss \(\mathcal{L}(\theta, \phi^\prime_i;\mathcal{D}^{spt}_i)\)\;
        Update user embedding \textbf{(UE adaptation)}:\;
        \(\phi^\prime_i \gets \phi^\prime_i - \alpha^\prime \nabla_{\phi^\prime_i} \mathcal{L}(\theta, \phi^\prime_i;\mathcal{D}^{spt}_i)\)\;
        \(\alpha^\prime \gets d \cdot \alpha^\prime\)\;
      }
      Compute regression loss \(\mathcal{L}(\theta, \phi^\prime_i;\mathcal{D}^{qry}_i)\)\;
    }
    Compute meta-loss \(\mathcal{L}^{meta}(\theta)\)\ with Eq.(2)\;
    Update shared parameters \textbf{(Meta-update)}:\;
    \(\theta \gets \theta - \beta \nabla_{\theta} \mathcal{L}^{meta}(\theta)\)\;
  }
  \caption{Meta-learning for Fast User Adaptation}
  \label{alg:01}
\end{algorithm}

\begin{figure*}[t]
  \centering
  \includegraphics[width=\linewidth]{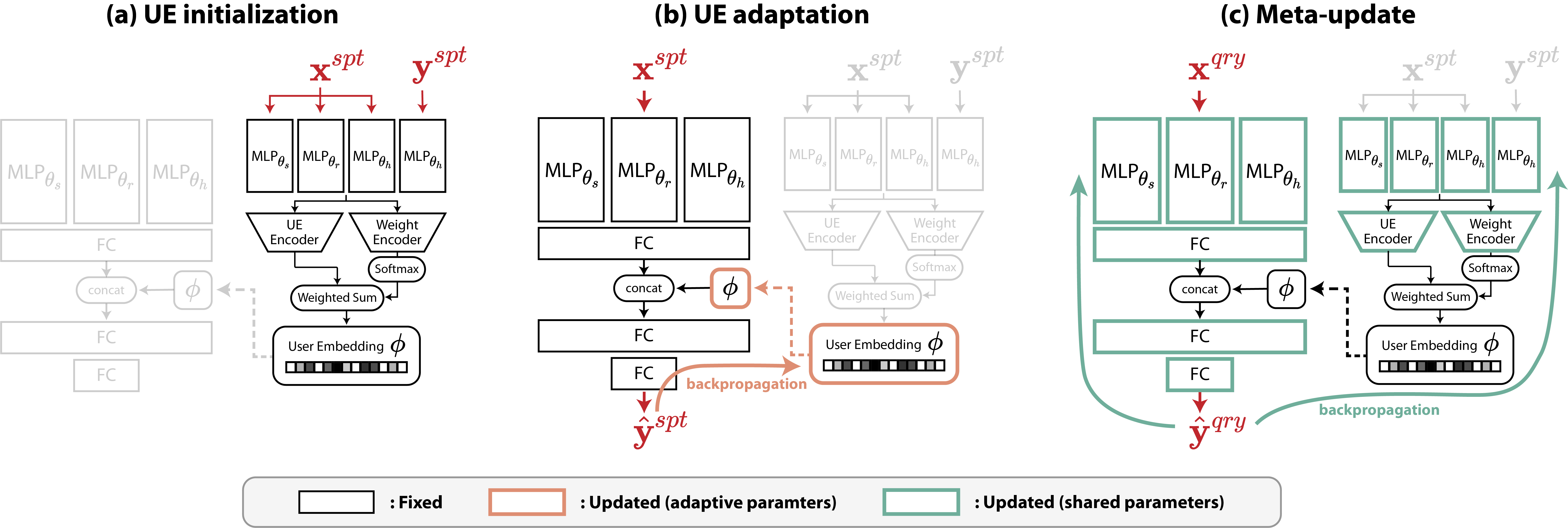}
  \caption{Three steps of the proposed meta-learning procedure for fast user adaptation: (a) UE initialization with small data samples, (b) UE adaptation with small data samples via a few gradient steps, and (c) meta-update of shared parameters between users.}
  \label{fig:03}
\end{figure*}

The meta-training process of the proposed model is realized in three steps: (1) UE initialization, (2) UE adaptation, and (3) meta-update, as summarized in Algorithm~\ref{alg:01} and Fig.~\ref{fig:03}. User-specific parameters (i.e., \(\phi\)) are determined via the UE initialization and UE adaptation steps, and the parameters shared across various users (i.e., \(\theta\)) are learned via the meta-update step.

During the first step, UE initialization, \(\phi_i\) is obtained by passing \(\mathcal{D}^{spt}_i\) through the initialization network described in Section~\ref{sec:3-B}. Subsequently, in the UE adaptation step, \(\phi_i\) is updated to \(\phi^\prime_i\) by backpropagating \(\mathcal{L}(\theta, \phi_i;\mathcal{D}^{spt}_i)\), which is the regression loss of the prediction network conditioned by \(\phi_i\) when using \(\mathcal{D}^{spt}_i\). Either one or a few gradient steps relative to \(\phi_i\) can be taken. For example, using one gradient step, \(\phi^\prime_i\) can be computed as follows:
\begin{equation}
  \phi^\prime_i = \phi_i - \alpha \nabla_{\phi_i} \mathcal{L}(\theta, \phi_i;\mathcal{D}^{spt}_i),
  \label{eq:01}
\end{equation}
where \(\alpha\) denotes the inner learning rate. In the case of adopting multiple gradient steps, decaying the inner learning rate by the decay rate \(d\) for every update can benefit a delicate adaptation.

In the meta-update step, we update \(\theta\) using \(\mathcal{D}^{qry}_i\), while fixing the \(\phi^\prime_i\) determined by \(\mathcal{D}^{spt}_i\). The meta-update is performed by overseeing the user-specific prediction results and values of the adapted \(\phi^\prime_i\) from multiple users. The meta-loss we propose consists of three terms expressed as:
\begin{equation}
  \begin{split}
    \mathcal{L}^{meta}(\theta) = &\frac{1}{K}\sum_i(\mathcal{L}(\theta, \phi^\prime_i;\mathcal{D}^{qry}_i) \\
    &+ \lambda_1\norm{stopgrad(\phi^\prime_i)-\phi_i}^2_2)\\
    &+ \lambda_2\norm[\bigg]{\frac{\sum_i\phi^\prime_i}{K}}^2_2,
    \label{eq:02}
  \end{split}
\end{equation}
where \(K\) represents the number of users sampled for the one meta-update, whereas \(\lambda_1\) and \(\lambda_2\) are the weights of each loss term. The first term aims to update \(\theta\) to reduce the regression loss of \(\mathcal{D}^{qry}_i\) in the prediction network with adapted user-specific parameters \(\phi^\prime_i\), which allows \(\theta\) to infer general human movements across the users. The second term, which is inspired by~\cite{rusu2019meta}, enables the initialization network to output an effective \(\phi_i\) close to the adapted \(\phi^\prime_i\). The \(stopgrad(\phi^\prime_i)\) denotes that we consider it as a constant, therefore the derivative of \(stopgrad(\phi^\prime_i)\) with respect to \(\theta\) is zero. The last term encourages the average of multiple \(\phi^\prime_i\), each adapted to a different user, to move toward a zero vector. The common nonzero bias from multiple \(\phi^\prime_i\) indicates the general movement tendency of the users. Therefore, by forcibly reducing the bias, we induce a general tendency to learn by \(\theta\). In addition, the \(\phi^\prime_i\) of different users are induced to be disentangled around zero and eventually learned to address individual differences. Taken together, \(\theta\) is updated with a gradient step of the meta-loss; therefore,
\begin{equation}
  \theta \gets \theta - \beta \nabla_\theta \mathcal{L}^{meta}(\theta),
  \label{eq:03}
\end{equation}
where \(\beta\) indicates the outer learning rate. The prediction and initialization networks were gradually trained by repeatedly performing the three steps using the sampled batch for each iteration. During the evaluation phase, the meta-update step is omitted, and only the UE initialization and UE adaptation processes are performed to achieve user-adapted motion prediction.

\section{Experiment and Results}

\subsection{Data Acquisition}
\label{sec:4-A}

To evaluate our meta-learning approach for the fast user adaptation problem, it is necessary to collect a multi-person motion dataset acquired in a situation wherein a human physically interacts with a robot. A representative pHRI situation occurs when a user performs a target task with \textit{haptic guidance} from a robot. The haptic guidance system, which has been recognized as a promising human--machine interface~\cite{abbink2012haptic}, can be defined as a system in which the control input of the target task is determined by the interaction between the force exerted by the user and the guiding force of the robot~\cite{abbink2010neuromuscular}. Because users are free to decide how much guiding force they will accept every moment, different users exhibit different responses to the guiding force. Therefore, it is possible to obtain a wide variety of human motion data, which is suitable for evaluating fast user adaptation performance.

\begin{figure}[t]
  \centering
  \includegraphics[width=\linewidth]{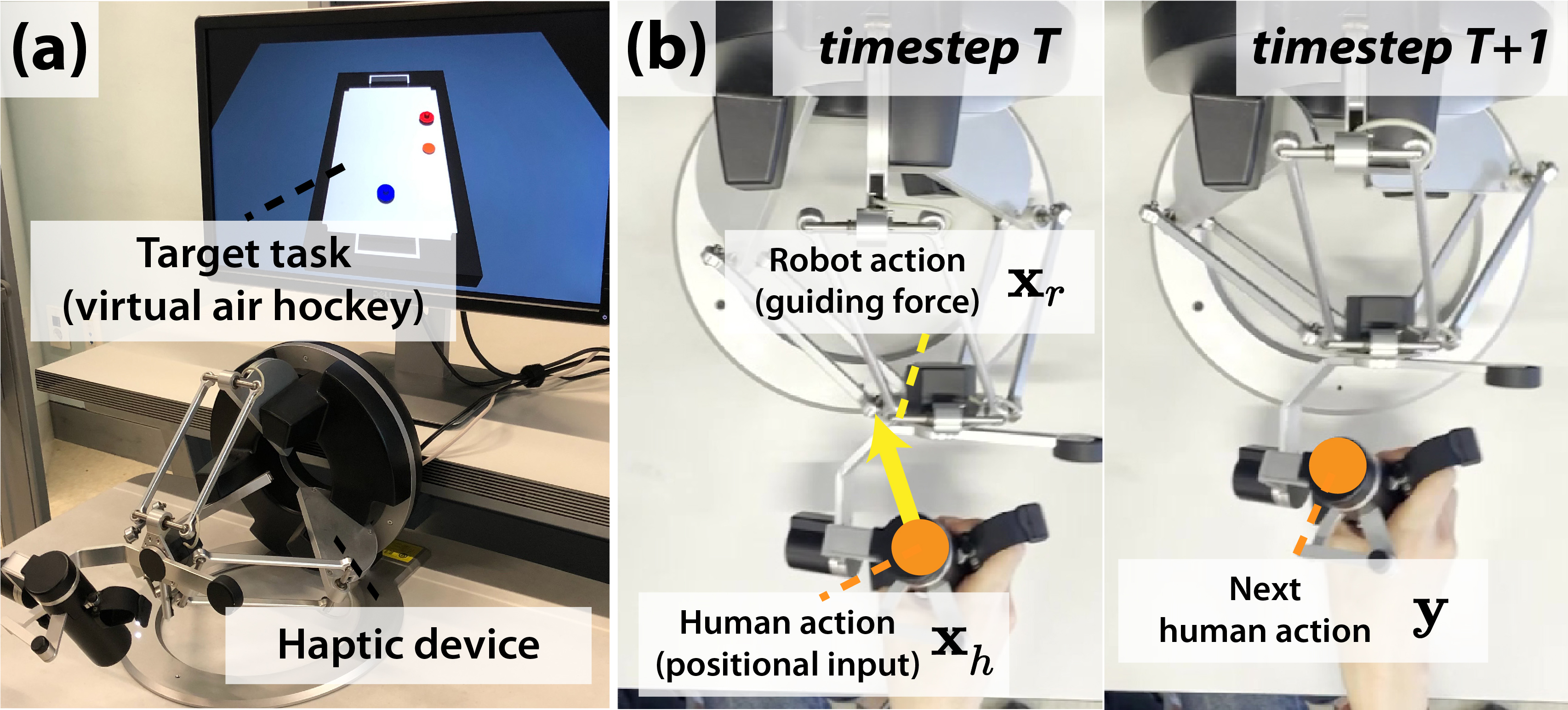}
  \caption{(a) Experimental setup for data collection. (b) Example of interaction process between a user and a robot on the haptic device.}
  \label{fig:04}
\end{figure}

We utilized a dataset consisting of motion data from 20 participants performing a target task with haptic guidance, which was collected in our previous work~\cite{moon2021optimal}. In the experimental environment, as illustrated in Fig.~\ref{fig:04}(a), the participants were instructed to play a virtual air hockey game controlled with a haptic device. In the hockey environment, a user receives points by smashing the puck with their paddle and putting the puck into the opponent's goalpost. Conversely, the user loses points if they fail to defend a puck heading to their goal. Through the haptic device, the participants consistently received a guiding force to assist them; however, they were allowed to choose whether to follow the guidance. The robotic guiding force dynamically changed according to the opponent's play.

The entire dataset is composed of the following three data types suitable for our prediction model structure, as described in Section~\ref{sec:3-B}. First, the human action data (corresponding to \(\textbf{x}_h\) and \(\textbf{y}\)) consist of 2-D position vectors of the end-effector of the haptic device determined by the user, which was transmitted to the target task as the control input. Second, the robot action data (corresponding to \(\textbf{x}_r\)) consist of 2-D force vectors that the robot exerts on the user. Finally, the state data of the target task (corresponding to \(\textbf{x}_s\)) consist of the 2-D position and velocity vectors of the paddles and the puck in the virtual air hockey environment. Fig.~\ref{fig:04}(b) presents an example of the interaction process between a user and a robot that occurred on a haptic device. In the timestep \(T\), the user action corresponds to the position vector \(\mathbf{x}_h\) of the end-effector of the haptic device (marked as an orange dot). Simultaneously, the user receives a guiding force \(\mathbf{x}_r\) (marked as a yellow arrow). The example illustrates the next user action (i.e., the position vector \(\mathbf{y}\) at timestep \(T+1\)) when the user follows the guiding force. If the guiding force does not match the user's intention, the user is allowed to control the end-effector in their desired direction by applying a force exceeding the guiding force.

A total of 1.52 M timesteps of data (i.e., 1.52 M pairs of \((\mathbf{x}, \mathbf{y})\)) were collected from 20 participants aged \(25.58 \pm 2.09\) (mean \(\pm\) standard deviation across participants) years. Each participant performed \(84.25 \pm 3.52\) trials, and the length of data collected per participant (i.e., \(N_i\)) was \(76.2 \pm 12.7\) K timesteps. The total duration of the data collection per participant was approximately 1--1.5 h, and the details of the procedure are described in~\cite{moon2021optimal}. All collected data were normalized. The mean values of each participant's action (i.e., \(\mathbf{y}\)) were \(-0.056 \pm 0.043\) (horizontal direction) and \(-0.152 \pm 0.053\) (vertical direction).

According to Article 15 (2) of the Bioethics and Safety Act and Article 13 of the Enforcement Rule of Bioethics and Safety Act in Korea, a research project ``which utilizes a measurement equipment with simple physical contact that does not cause any physical change in the subject'' (Korean to English translation by the authors) is exempted from the approval. The entire experimental procedure was designed to use only a haptic device and a monitor that did not cause any physical changes in the subject.

\subsection{Experimental Setting}

\textbf{Evaluation:} To evaluate the fast user adaptation performance, we measured the prediction performance for \(\mathcal{M}^{test}\), which consisted of user datasets that were not used for training, that is, users in \(\mathcal{M}^{test}\) were not included in \(\mathcal{M}^{tr}\). We assumed a realistic sampling situation during the evaluation procedure. If \(\mathcal{D}^{spt}_i\) (i.e., batch for adaptation) and \(\mathcal{D}^{qry}_i\) (i.e., batch for prediction) were randomly sampled from the entire user dataset \(\mathcal{U}_i\) as in the training procedure (Algorithm~\ref{alg:01}), there may be unrealistic cases of predicting the current human motion while utilizing the later motion data within the same episode for adaptation. To prevent this, for the evaluation, we utilized data from different episodes to adapt the model and to validate the prediction, that is, \(\mathcal{D}^{spt}_i\) and \(\mathcal{D}^{qry}_i\) consisted of data from separate episodes.

As a metric for prediction performance, we adopted the mean squared error (MSE) between the predicted value and the ground-truth value of the next user action. Five-fold cross-validation was conducted to reduce the effect of the discrepancy between the training and test datasets. Therefore, the 20-user dataset was divided into five sub-datasets consisting of four users each, and a total of five training-validation processes were performed using five \((\mathcal{M}^{tr}, \mathcal{M}^{test})\) pairs. The averaged values of the resulting five prediction errors (MSE) were used to compare the learning methods.

\textbf{Baselines:} We set baseline methods, including non-meta-learning and meta-learning approaches. Ahead of both approaches, the zero-velocity baseline~\cite{martinez2017human} was adopted, which assumed that the user maintained the previous action. This helps to understand the prediction performance of the other learning methods at an appropriate scale.

As a non-meta-learning approach, we trained the same structured prediction model using a conventional supervised learning method. For a fair comparison with the adaptation process of the meta-learning approach, we aimed to determine the extent to which performance is improved when the model trained with a non-meta-learning method goes through the parameter update process with \(\mathcal{D}^{spt}\) (i.e., fine-tuned). Therefore, we implemented the non-meta-learning baselines in two ways: when the model parameters were fixed and when they were fine-tuned with a few gradient steps, similar to the adaptation steps of the meta-learning approach.

For the meta-learning approaches, we evaluated the most representative methods, MAML and Reptile.
In addition, we tested the performance of integrating the model regression network (MRN)~\cite{wang2016learning} into the adaptation process of MAML, which is equivalent to PAML~\cite{gui2018few}.
Our approach to solve the fast user adaptation problem was to separate the adaptive parameters and enforce them to represent only user-specific movements. Therefore, we also considered two other meta-learning baselines, ANIL and CAVIA, that distinguish between adaptive and shared parameters. There is a difference in how each method divides the parameters. Whereas ANIL has the body layers be fixed and the last layer adaptive, CAVIA adopts the adaptive parameters that join as an auxiliary input for the body layers.

\textbf{Implementation details:} All learning-based baseline methods were implemented using our prediction network structure (Fig.~\ref{fig:02}), except for the user embedding \(\phi\). Only the CAVIA method adopts additional adaptive parameters to the body layers; therefore, it can be implemented using the same structure as our prediction network. We set the hyper-parameters of our method and baseline methods to be as similar as possible. For the meta-learning approaches, including our method, the adaptation process was conducted through five gradient steps based on a stochastic gradient descent optimizer. CAVIA and the proposed method that updates \(\phi\) (we set the size of \(\phi\) to 32) adopted an inner learning rate of 0.05, and the other methods, which directly update the parameters of the network layers, employed an inner learning rate of 0.01, for both training and evaluation phases. Exceptionally, in the training phase of the MAML and ANIL algorithms, a small inner learning rate of 0.003 was adopted, because it exhibited more stable learning. Regarding the meta-update, an Adam optimizer with an outer learning rate of 0.001 was adopted for all methods, and each model was trained for 500 K steps. The batch sizes of \(\mathcal{D}^{spt}\) and \(\mathcal{D}^{qry}\) were set to 1 K samples. For the methods adopting multi-person data for one meta-update, such as Reptile or our method, the total number of data samples used for one meta-update was maintained at 1 K samples by adopting 200 samples each from five different users. Regarding the non-meta-learning approaches, an Adam optimizer with a learning rate of 0.001 was employed to train the model for 500 K steps, using batches consisting of 1 K samples. In the fine-tuned baseline case, five gradient steps with a learning rate of 0.01 were taken in the same manner as the adaptation process in the meta-learning approaches.

\subsection{Results}

\begin{table}[t]
  \caption{MSE Comparison of Different User Prediction Methods}
  \label{tab:01}
  \begin{center}
  \begingroup
  \setlength{\tabcolsep}{10pt} 
  \renewcommand{\arraystretch}{1.28} 
  \begin{tabular}{c|c||c}
    \hline
    \multicolumn{2}{c||}{\textbf{Methods}} & \textbf{MSE}\\
    \hline
    \multicolumn{2}{c||}{Zero-velocity~\cite{martinez2017human}} & $3.127\mathrm{e}{-3}$ \\ \hline
    \multirow{2}{*}{Non-meta-learning} & Fixed & $1.295\mathrm{e}{-5}$\\ \cline{2-3}
    & Fine-tuned & $1.268\mathrm{e}{-5}$\\ \hline
    \multirow{4}{*}{Meta-learning} & MAML~\cite{finn2017model} & $1.194\mathrm{e}{-5}$ \\ \cline{2-3}
    & Reptile~\cite{nichol2018first} & $1.175\mathrm{e}{-5}$\\ \cline{2-3}
    & MAML + MRN~\cite{wang2016learning} & $1.188\mathrm{e}{-5}$\\ \cline{2-3}
    & ANIL~\cite{raghu2019rapid} & $1.221\mathrm{e}{-5}$\\ \cline{2-3}
    & CAVIA~\cite{zintgraf2019fast} & $1.115\mathrm{e}{-5}$\\ \hline
    \multirow{2}{*}{Ablation} & {w/o UE initialization} & {$1.064\mathrm{e}{-5}$}\\ \cline{2-3}
    & {w/o UE bias reduction} & {$1.090\mathrm{e}{-5}$}\\ \hline
    \multicolumn{2}{c||}{\textbf{Our method}} & $\mathbf{1.045\mathrm{e}{-5}}$\\ \hline
  \end{tabular}
  \end{center}
  \endgroup
\end{table}

The quantitative results are presented in Table~\ref{tab:01}. Compared to the zero-velocity baseline, all learning-based baselines exhibited a significantly lower prediction error. Meta-learning approaches exhibited more accurate performance than non-meta-learning approaches. The fine-tuned baseline did not exhibit a significant difference in prediction performance when compared to the fixed baseline. In contrast, the superior performance of the meta-learning approaches indicates that they succeeded in adapting rapidly to predict the movements of previously unseen users with only a few gradient steps. In other words, a meta-learning approach can be an effective solution for fast user adaptation problems.

A performance difference existed within the meta-learning baselines, and CAVIA exhibited the best performance among the baselines. As stated in~\cite{zintgraf2019fast}, CAVIA outperformed MAML in solving various problems, such as image classification or reinforcement learning, and we observed the same results for the fast user adaptation problem. Notably, CAVIA and ANIL both applied separate parameters for adaptation; however, the performance of ANIL did not differ from that of MAML, whereas CAVIA exhibited better performance. This implies that designing the stage at which the model divides shared and adaptive parameters plays an important role in improving user adaptation performance. For example, in ANIL, because all body layers are fixed across the users and only the last layer is adapted, shared features are obtained from input data and user-specific computation is performed in the process of assembling the shared features. However, in CAVIA, because the adaptive parameters join in the middle stage, the model can consider both the shared and user-specific features, which is consistent with our approach that considers user motion as a combination of general motion across users and motion with individual differences.

Our method outperformed all other baseline methods. In particular, the superior performance over CAVIA indicates the benefits of the two components we proposed: (i) the user embedding initialization and (ii) the meta-update reducing the non-zero bias of multiple user embeddings. We verified the contribution of each component by conducting an ablation study. We implemented the prediction models by excluding each component as follows:
\begin{itemize}
  \item \textit{Without UE initialization:} Each user embedding was initialized with a zero vector. The meta-loss in (\ref{eq:02}) without the second term was utilized.
  \item \textit{Without UE bias reduction:} Each meta-update was performed with a batch sampled from a single user. The meta-loss in (\ref{eq:02}) without the last term was utilized.
\end{itemize}
As shown in Table~\ref{tab:01}, both models without either UE initialization or UE bias reduction outperformed all the baselines, indicating that each component contributed to the performance improvement of the proposed method with UE initialization and UE bias reduction. Among the two components, the bias reduction of user embeddings contributed more than the UE initialization.

\subsection{Analysis of User Embeddings}

\begin{figure}[t]
  \centering
  \includegraphics[width=\linewidth]{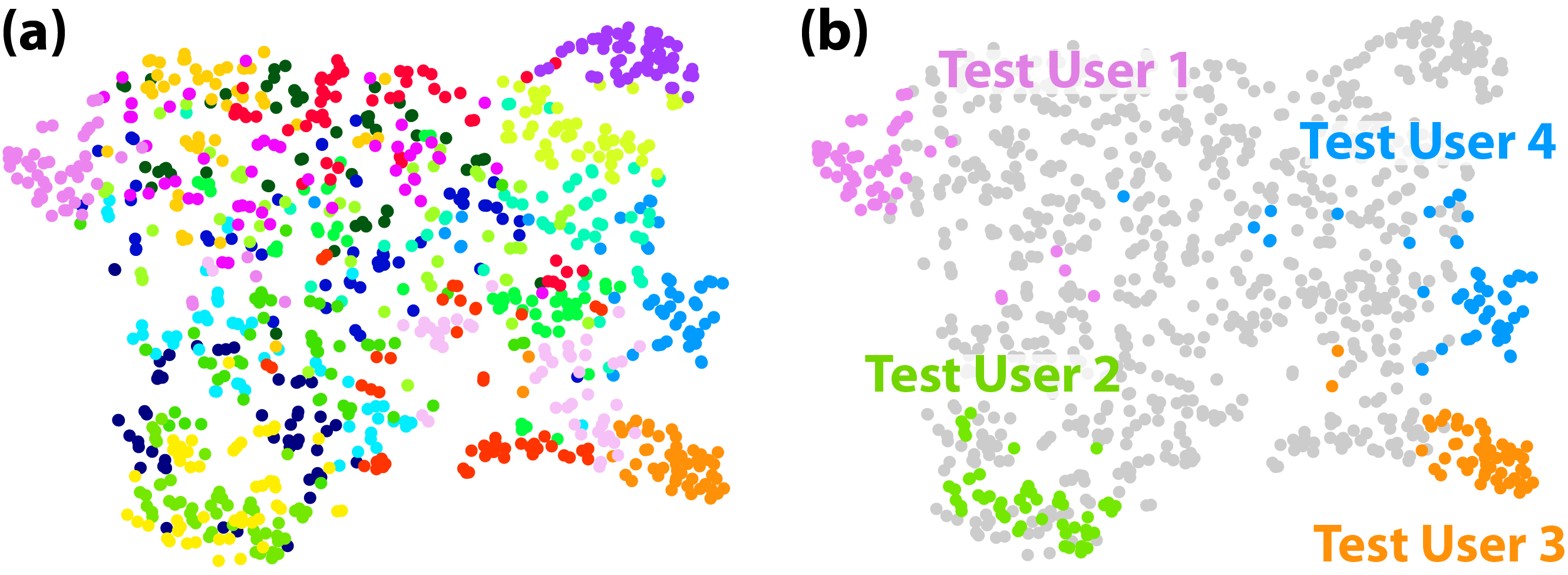}
  \caption{(a) t-SNE plot of the user embeddings adapted for 20 users based on our method. Different colors represent different users. (b) Same as (a), but highlighting the embeddings of four unseen users.}
  \label{fig:05}
\end{figure}

Implementing the adaptation with independent embedding parameters, rather than updating the entire model parameters such as MAML, has the advantage of being easily interpretable~\cite{zintgraf2019fast}. Moreover, the embedding can efficiently reflect the behavioral characteristics of each user in a low-dimensional space because the embeddings are induced to exclusively learn the individual differences.

We qualitatively investigated the user adaptation performance of our method by visualizing user embeddings adapted to different users. Within the dataset of 20 participants, we sampled 50 batches, each comprising 1 K timesteps of data, for each user. Using our trained model, one user embedding per batch was produced via UE initialization and adaptation processes (i.e., five gradient steps). Fig.~\ref{fig:05}(a) presents the t-SNE~\cite{van2008visualizing} projection results of the 32-dimensional user embeddings onto 2-D space. It can be observed that user embeddings generated from the same user dataset (i.e., same-colored dots) are clustered together, and user embeddings from different user datasets are disentangled. This indicates that our user adaptation approach succeeded in rapidly extracting user characteristics using only small data samples. Furthermore, Fig.~\ref{fig:05}(b) highlights the embedding generation results of four users whose data were not employed for model training. Our method effectively responded to previously unseen users, as evidenced by the well-disentangled embeddings of test users.

\section{Discussion and Conclusion}

We focused on the one-step prediction of human behavior because we considered a situation in which a human was guided by a robot at every timestep, and the robotic guidance was dynamically changing and unpredictable. However, our proposed meta-learning algorithm can be applied to train a model that predicts human motion over a longer time horizon (assuming that the robotic guidance over the horizon is known in advance). In this case, the prediction network (Fig.~\ref{fig:02}) needs to be modified based on recurrent neural networks instead of the current fully connected layers~\cite{fragkiadaki2015recurrent, gui2018few}. The modified model can be trained using the same meta-learning procedure as in Algorithm~\ref{alg:01}.

In this study, we introduced a meta-learning approach to train a human prediction model that facilitates fast user adaptation, that is, allowing the model to swiftly respond to previously unseen users with small data samples. We proposed a meta-learning algorithm and a model structure that predicts the movement of individual users in pHRI situations. The superior prediction performance of the proposed method was quantitatively verified using a 20-user dataset. We also qualitatively validated the fast user adaptation performance of the proposed method by investigating the disentanglement of user embeddings adapted to various users. The proposed meta-learning framework for fast user adaptation can be useful when robots cannot obtain sufficient data from new users, such as service robots that encounter many people in a short period of time.



\bibliographystyle{IEEEtran.bst}
\bibliography{references.bib}
\end{document}